\documentclass[journal]{IEEEtran}
\usepackage{amsmath}
\usepackage{amsthm}
\usepackage{bm}
\usepackage{algorithm}
\usepackage{algpseudocode}
\usepackage{graphicx}
\usepackage{color}
\usepackage{natbib}
\usepackage{hyperref}

\begin{document}
\title{Large-scale image segmentation based on distributed clustering algorithms}
\author{Ran Lu, Aleksandar Zlateski and H. Sebastian Seung
\thanks{Ran Lu is with the Neuroscience Institute, Princeton University, Princeton, NJ 08544 USA (e-mail: ranl@princeton.edu).}
\thanks{Aleksandar Zlateski is with FAIR, New York NY, 10007 USA (aleksandar.zlateski@gmail.com). Work was performed while author was with MIT.}
\thanks{H. Sebastian Seung is with the Dept. of Computer Science and the  Neuroscience Institute, Princeton University, Princeton, NJ 08544 USA. (e-mail: sseung@princeton.edu).}}
\maketitle
\begin{abstract}
    Many approaches to 3D image segmentation are based on hierarchical clustering of supervoxels into image regions. Here we describe a distributed algorithm capable of handling a tremendous number of supervoxels. The algorithm works recursively, the regions are divided into chunks that are processed independently in parallel by multiple workers. At each round of the recursive procedure, the chunk size in all dimensions are doubled until a single chunk encompasses the entire image. The final result is provably independent of the chunking scheme, and the same as if the entire image were processed without division into chunks. This is nontrivial because a pair of adjacent regions is scored by some statistical property (e.g. mean or median) of the affinities at the interface, and the interface may extend over arbitrarily many chunks. The trick is to delay merge decisions for regions that touch chunk boundaries, and only complete them in a later round after the regions are fully contained within a chunk. We demonstrate the algorithm by clustering an affinity graph with over 1.5 trillion edges between 135 billion supervoxels derived from a 3D electron microscopic brain image.
\end{abstract}

\section{Introduction}
Reconstructing neural connectivity from electron microscopic (EM) images has become an important method for neuroscience. In light of advancing technologies, the size and quality of image stacks collected have been growing exponentially in recent years. High quality segmentations of these images are essential to extract precise information of neural circuits. However, neurons have complex morphology and segmenting them from grey scale EM images is a challenging task.

Convolutional networks are now standard for EM image segmentation \citep{lee2019convolutional}. Here we focus on one of these approaches \citep{Dorkenwald2019.12.29.890319}: First we use convolutional networks to calculate inter-voxel affinities between nearest neighbors. The affinities are then converted to segmentation by classical algorithms including watershed and agglomeration. Variants of this approach include adding affinities between more distant voxels as an auxiliary task during training \citep{DBLP_abs-1909-09872}, improved training techniques \citep{8364622} etc.

After the affinities are determined, the image is segmented using classical algorithms that do not depend on machine learning. The convolutional networks only predict the affinities within their fields of view (FOV). For large objects that cannot be included in the FOV, watershed and agglomeration algorithm extend the affinities into the entire image stack to create complete segments. If the models' predictions were perfect, this could be done by calculating the connected components over the affinities, however in reality the errors and noises in the output of the convolutional networks are unavoidable. Using sophisticated segmentation algorithms \citep{Wolf_2018_ECCV,multicut} helps achieve higher quality by suppressing these errors through statistical means. For this purpose, most of the popular agglomeration algorithms take an iterative approach, agglomerate fragments in descending order of the affinities between them, and update the affinities through statistical methods after each agglomeration. This introduces a global dependency on the input data. A local change in the input can cascade through the iteration process and induce errors in a far region of the output. Thus, it is common to process the entire input together. Currently, most of the proposed methods were demonstrated on small image stacks that completely fit into the working memory of a single machine. When segmenting larger images, the input images are divided into (overlapping) chunks and processed independently; After all chunks are segmented they are stitched together by heuristics \citep{8364622}. As aforementioned algorithms have global dependencies, this per-chunk processing could introduce additional errors.

Here, we propose a novel approach to segmenting large image stacks. The input images are similarly divided into chunks so they can be distributed in computer clusters or cloud computing services and processed in parallel. However, the algorithm we use to agglomerate within each chunk and stitch chunks together guarantees that no errors are introduced throughout the process. Our method generalizes to a large class of hierarchical clustering algorithms -- as long as affinities between clusters are updated by linkage criteria that satisfy the reducibility condition \citep{2011arXiv1109.2378M}. Using notations we will explain in more details in the following sections, considering three clusters $I, J$ and $K$, this is true for any linkage criterion satisfying:
\begin{equation}
    %\min \left( A\left( I, K \right), A\left( J, K \right) \right) \le A \left( I\cup J, K \right) \le \max \left( A\left( I, K \right), A\left( J, K \right) \right)
    A \left( I\cup J, K \right) \le \max \left( A\left( I, K \right), A\left( J, K \right) \right)
    \label{eq:mean}
\end{equation}
where $A\left( I, J \right)$ is the affinity between the clusters $I$ and $J$. Many popular linkage criteria used in EM image segmentation fall into this category. This is not a surprise, since the input of these clustering algorithms are affinities predicted by convolutional networks. These affinities can be interpreted as estimated probabilities of two regions belonging to a single segment; updating them by some statistical averaging is natural.

It is well known in the literature, when the linkage criteria satisfy the reducibility condition, some agglomeration operations can be performed out of order. In this paper we point out the reducibility condition also allow algorithms to perform agglomeration within partial input and avoid introducing errors due to the incomplete local information. Our algorithm treats the supervoxels touching the chunk boundary in a special manner; we use this information to avoid fictitious mergers or splits. After we merge all the clusters that can be agglomerated, we are left with an intermediate, partial, segmentation containing the residual clusters that cannot be safely agglomerated within the chunk, as some non--local information might affect the decision of whether to merge two clusters or not. We suggest an elegant solution to merge the partially segmented chunks: we simply collect the residual clusters and apply the same algorithm on them, now with no boundaries. More generally we can start from small chunks, and gradually merge them into bigger and bigger chunks, during which we applying the same agglomeration algorithm at each step, and updating the boundaries accordingly. Once we merge all chunks together,  we obtain a segmentation of the entire input image stack. Using our method, all the segments are agglomerated by the same algorithm. There is no need to develop a separate stitching method. It not only simplifies the segmentation pipeline, also eliminates errors introduced by the stitching. In fact, it can be proven that the segmentation obtained this way is independent of the chunking scheme, thus identical to the segmentation created by a single machine that has all the global information.

\section{Segment images with hierarchical agglomerative clustering algorithms}
Image segmentation classifies voxels into domains representing different objects in the input image, it can be treated as a clustering problem. The input is an affinity graph $G_0 = \left( V_0, E_0, A_0 \right)$. The nodes in $V_0$ represent single voxel clusters, $E_0 \subset V_0\times V_0$ contains edges between the nodes, and the map $A_0: E_0 \rightarrow \mathcal{R}$ stores the affinities associated with the edges. For image segmentation, there is a natural definition of locality. Following this intuition, we often only consider the correlations between direct neighboring voxels. This makes the affinity graph sparsely connected. For example if we only consider correlations between voxels and their direct neighbors in of a 3D image stack, $|E_{0}| \approx 3 |V_{0}|$. Locality is a crucial assumption to our distributed algorithm: if voxels can have arbitrary correlations at any distance, we must consider the entire $G_0$ when clustering voxels. This assumption can be extended to include long range correlations studied in \citep{DBLP_LeeZLJS17,Wolf_2018_ECCV}, as long as the long range correlations are still confined within some bounded FOV. Throughout the agglomeration, we track the process with a region adjacency graph (RAG) $G = \left( V, E, A \right)$. At the beginning the RAG is simiply the affinity graph $G=G_0$. As we agglomerate the voxels together, we update $G$ by merging nodes and updating the weights between them according to the linkage criterion. In the end we collect the edges corresponding to the clustering operations to construct the dendrogram.

Various linkage criteria have been invented for clustering problems in different fields. For image segmentation, only consider affinities between direct neighboring voxels, we can use the affinities between the boundary voxels to determine the affinities between two clusters.\footnote{If we are interested in long range affinities we have to include the long range edges across the boundary as well}. In this case linkage criteria are often some statistical property of the boundary voxels' affinity distribution: max, mean, quantiles, etc. This can be extended to incorporate extra input based on the clusters' intrinsic properties. In \citep{DBLP_ZlateskiS15} the max affinity linkage criterion was augmented by size thresholds. One can also include prior knowledge of the underlying image stack to improve the segmentation. For example, assigning negativing affinities to voxel pairs with different semantic labels to prevent false mergers or reweighting the affinities to account for varying image quality.

There are a few more considerations specific to image segmentation we want to mention. First, our goal is to cluster voxels into domains representing the objects we want to study. There is little interest to agglomerate everything into a single cluster and obtain a complete dendrogram. For image segmentation, it is sufficient to stop clustering at a threshold $T$ and output the truncated dendrogram or a flat segmentation. Also, as we alluded in the Introduction section, commonly we start clustering from supervoxels\footnote{Small clusters of voxels that are likely belong to the same object}. Starting from single voxels can be considered as a special case. These supervoxels can be created by faster but less robust algorithms like watershed or obtained from an existing (over)segmentation. The affinities between supervoxels are readily defined by the linkage criterion. Using supervoxels can reduce the size of the input RAG and simplify the computation. It also allows us to perform multi-stage agglomeration by switching algorithm and parameters, treating the segments of the previous step as input supervoxels to optimize the final results. Finally, since there are no intrinsic differences, we will use clusters, supervoxels, segments as interchangeable terms.

\section{Chunk-based agglomerative clustering algorithm}
To explain our chunk-based clustering algorithm, we start by reviewing a generic algorithm similar to the one studied in \citep{article}. The input data is a RAG $G\left( V, E, A \right)$. We also supply a threshold $T$ and stop the agglomeration after the affinities of all edges are lower than it. Before starting the agglomeration, we create a heap $H$ from the edge list $E$, which are used to extract the edge with the highest affinity to perform the next clustering operations. The output of the agglomeration is a dendrogram $D$

\begin{algorithm}
    \caption{Generic clustering algorithm}\label{algorithm:generic}
    \begin{algorithmic}[1]
        \Procedure{$\rm{AGGLOMERATE\_GENERIC}$}{$G, T$}
        \State $H \gets$ Heap from $G$
        \State $D \gets \emptyset$
        \While{$H$ is not empty}
        \State $\left\{ u, v \right\} \gets$ top element with the maximal affinity in $H$\label{line:top_edge}
        \State pop $\left\{ u,v \right\}$ out of $H$
        \If {$G.A[\left\{ u, v \right\}] < T$}
            \State \textbf{break}
        \EndIf
        \State $D \gets \left\{ \left( \left\{ u, v \right\}, G.A\left[ \left\{ u,v \right\} \right] \right) \right\} \cup D$
        \For{$w \in G.Adj[v]$}
        \Comment{Update $G$: merging $v$ with $u$}
        \If{$w \in G.Adj[u]$}
        \State merge $\left\{ w, v \right\}$ with $\left\{ w, u \right\}$
        \State pop $\left\{ w, v \right\}$ out of $H$
        \Else
        \State assign $\left\{ w, v \right\}$ to $\left\{ w, u \right\}$
        \EndIf
        \EndFor
        \EndWhile
        \State \textbf{return} $D$
        \EndProcedure
    \end{algorithmic}
\end{algorithm}

The global nature of Algorithm \ref{algorithm:generic} is obvious in line \ref{line:top_edge}: We extract the edge with the highest affinity from the heap and agglomerate the two nodes connected by it. If the input RAG is incomplete, we will likely agglomerate a different set of edges, and the output dendrogram $D$ will be different.

Fortunately, there are known examples of linkage criteria allowing algorithms based on local properties. Most notably, when the linkage criteria satisfy the reducibility condition, one can prove that the same dendrogram can be constructed by agglomerating mutual nearest neighbors with arbitrary orders.\citep{1985mca..book.....M,2011arXiv1109.2378M,SAD_1977__2_3_24_0,CAD_1980__5_2_135_0}.

Consider three segments $I, J$ and $K$, and the affinities between them are represented as $A\left( I, J \right), A\left( I, K \right)$ and $A\left( J, K \right)$, reducibility condition requires:
\begin{eqnarray}
   && A\left( I, J \right) \ge \max\left( A\left( I, K \right), A\left( J, K \right) \right) \nonumber \\
   &\Longrightarrow& A\left( I \cup J, K \right) \le \max\left( A\left( I, K \right), A\left( J, K \right) \right)
    \label{eq:reducibility}
\end{eqnarray}
Intuitively, this condition indicates there is \emph{no reversal}. The affinity between the merged cluster $I\cup J$ and $K$ cannot be higher than both $A\left( I, K \right)$ and $A\left( J, K \right)$.

We call a segment $J$ the nearest neighbor of another segment $I$, if the affinity $A\left( I, J \right)$ satisfy:
\begin{equation}
    A\left( I, J \right) = \max\left( \left\{ A\left( I, K \right)\; \forall K \in V \right\} \right)
    \label{eq:nearest}
\end{equation}
Unless completely isolated, there is a nearest neighbor for each segment. Segment $I$ and $J$ are called mutual nearest neighbors when $I$ is also $J$'s nearest neighbor. We call the edge between $I$ and $J$ is a local maximal edge. One can readily see when the linkage criterion satisfy the reducibility condition, mutual nearest neighbors always merge with one another. Agglomerating the rest of the RAG cannot create an edge with higher affinity associated with $I$ or $J$ because the reducibility condition forbids such reversal. Similarly, agglomerating $I$ and $J$ together will not create edges with affinities higher than $A\left( I, J \right)$. We can agglomerate $I$ and $J$ out of the standard order, the original dendrogram can be restore by a simple permutation. One can extend this reasoning to a rigorous proof using induction \citep{2011arXiv1109.2378M}.

Reducibility condition is famous for allowing $O( |V|^2)$ clustering algorithms like the nearest-neighbor chain \citep{CAD_1982__7_2_209_0,CAD_1982__7_2_219_0}, which have been studied extensively in the literature. We point out that for image segmentation, combining with the locality assumption, reducibility conditiona allow us to identify and agglomerate local maximal edges with partial inputs. The modified algorithm suited for chunked input based on this idea is presented in Algorithm \ref{algorithm:chunked}.

\begin{algorithm}
    \caption{Clustering algorithm with chunked input}\label{algorithm:chunked}
    \begin{algorithmic}[1]
        \Procedure{\rm{AGGLOMERATE\_CHUNK}}{$G^C, B^C, T$}
        \State $H \gets$ Priority queue from $G^C$
        \State $F \gets B^C$
        \State $D^C \gets \emptyset$
        \State $G^C_{F} \gets$ empty graph
        \While{$H$ is not empty}
        \State $\left\{ u, v \right\} \gets$ top element with the maximal affinity in $H$\label{line:lme_candidate}
        \State pop $\left\{ u,v \right\}$ out of $H$
        \If {$u \in F$ or $v \in F$} \label{line:freeze}
        \Comment{Freeze nodes and edges}
            \State $F \gets \left\{ u \right\} \cup F$
            \State $F \gets \left\{ v \right\} \cup F$
            \State add edge $\left\{ u, v \right\}$ into $G_{F}$
            \State \textbf{continue}
        \EndIf
        \If {$G^C.A[\left\{ u, v \right\}] < T$}
            \State \textbf{continue}
        \EndIf
        \State $D^C \gets \left\{ \left( \left\{ u, v \right\}, G^C.A\left[ \left\{ u,v \right\} \right] \right) \right\} \cup D^C$
        \For{$w \in G^C.Adj[v]$}
        \Comment{Update $G^C$: merging $v$ with $u$}
        \If{$w \in G^C.Adj[u]$}
        \State merge $\left\{ w, v \right\}$ with $\left\{ w, u \right\}$
        \State pop $\left\{ w, v \right\}$ out of $H$
        \Else
        \State assign $\left\{ w, v \right\}$ to $\left\{ w, u \right\}$
        \EndIf
        \EndFor
        \EndWhile
        \State \textbf{return} $D^C, G^C_F$
        \EndProcedure
    \end{algorithmic}
\end{algorithm}

In algorithm \ref{algorithm:chunked}, we change our notations of the input and output from $G$, $D$ to $G^C$, $D^C$ to emphasis we intend to apply it to a chunk. To make the new algorithm aware of the chunked input, we include an extra input $B^C$. It contains all the boundary supervoxels touching the artificial chunk boundaries\footnote{not to be confused with the dataset's real boundaries}. These supervoxels may be split among chunks, and the edges associated with them in $G^C$ may be incomplete. Because of these boundary supervoxels, some of the segments cannot be agglomerated within the chunk. We call these segments ``frozen segments'' and keep track of them in a set $F$. At the beginning, $F = B^C$. The edge $\left\{ u,v \right\}$ we extracted in line \ref{line:lme_candidate} by definition is the edge with the highest affinity in $G^C$. if neither $u$ nor $v$ is in $F$, $u$ and $v$ are mutual nearest neighbors and we can agglomerate them immediately. Otherwise it means whether we can agglomerate $u$ and $v$ depends on agglomeration decisions involving segments in $F$. We cannot merge $u$ and $v$ before we resolve those pending agglomerations. We add $u$ and $v$ to $F$ to forbid agglomeration involving $u$ and $v$ in future iterations\footnote{For readers familiar with nearest-neighbor chain algorithms, we freeze nearest-neighbor chains truncated by the chunk boundaries}. The frozen edges are saved into a graph $G^C_F$. Later we process them correctly after we stitch the chunks together.

The rest of the changes are straightforward: we can no longer discard all the edges below the agglomeration threshold $T$, because some of them are frozen edges belonging to $G^C_F$. In the end, we return the partial dendrogram, and the frozen edges for later steps.

\begin{figure}[htpb]
    \centering
    \includegraphics[width=7cm]{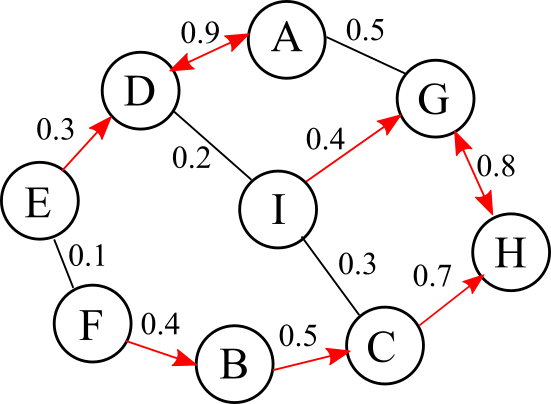}
    \caption{Mutual nearest neighbors and the nearest-neighbor chain in a RAG. We draw a red line between a segment and its nearest neighbor, with an arrow pointing from it to its nearest neighbor. For mutual nearest neighbors, we draw a red line with arrows in both directions. The numbers on top of the lines are the affinities between the segments. When we follow the arrows, we will eventually find a pair of mutual nearest neighbors, and the segments we visited are called a nearest-neighbor chain. In this graph, segment $A$ and $D$, segment $G$ and $H$ are the mutual nearest neighbors. It is straightforward to verify that we can agglomerate $G$ and $H$ first without altering the dendrogram.}
    \label{fig:chains}
\end{figure}

\begin{figure}[htpb]
    \centering
    \includegraphics[width=7cm]{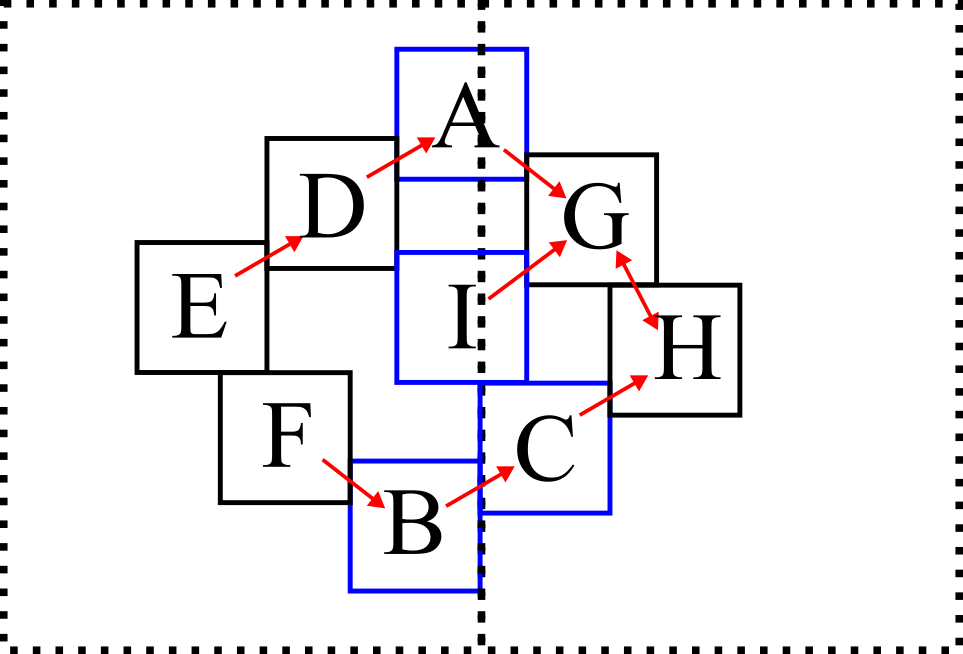}
    \caption{A RAG cut into two chunks. The graph is the same as the one we showed in Figure \ref{fig:chains}. Here we draw segments in squares and use the contact surfaces between them to represent the edges. The dashed lines indicate the chunk boundaries. The segments $A$, $B$, $C$, and $I$ in blue squares are boundary supervoxels. When we agglomerate segments within each chunk, we can not merge segments $A$ and $D$, because $A$ is a boundary supervoxels and we cannot see all its neighbors in either chunk. On the other hand, we can merge segments $G$ and $H$ within the right chunk. After we stitch the two chunks, the segments left can be clustered in the right order.}
    \label{fig:chunks}
\end{figure}

\section{Distributed clustering algorithm}
Algorithm \ref{algorithm:chunked} is the corner stone of our distributed clustering algorithm. It not only can agglomerate chunked input, but also can stitch those chunked results together to create segmentation of the entire dataset. For each chunk we get a partial dendrogram $D^C$, and a fronzen graph $G^C_F$. For simplicity we assume the chunks are not overlapping. In this situation, the segments across the chunk boundaries are avoided by Algorithm \ref{algorithm:chunked}, stitching them into a intermediate segmentation is trivial. We can merge all the $G^C_F$ together into a ``residual'' graph $G_{\rm res}$ representing the RAG of the intermediate segmentation by matching the segments and merging the incomplete edges. The edges remain in $G_{\rm res}$ are edges cannot be safely agglomerated within individual chunks. Now the RAG are extened to the entire dataset in $G_{\rm res}$, we can apply algorithm \ref{algorithm:chunked} again, with $B$ an empty set. The result is a dendrogram $D_{\rm res}$. Combine it with partial dendrograms $D^C$s we recover the dendrogram of the entire input image stack.

The two-step example above can be extended to a generic recursive algorithm \ref{algorithm:dist}. The input is a RAG $G$, and a boundary segment set $B$. If the chunk is small, we use \rm{AGGLOMERATE\_CHUNK} described in algorithm \ref{algorithm:chunked} to agglomerate it directly. If the input is too big, we call a \rm{DIVIDE\_CHUNK} function to cut it into smaller chunks. The \rm{DIVIDE\_CHUNK} function returns a collection of subchunks, each including its edge list $G^C$ and boundary supervoxels $B^C$. We then pass the subchunks to algorithm \ref{algorithm:dist} again to have them agglomerated. In the end, we collect the frozen edges in each subchunk and use the function \rm{COMBINE\_EDGES} to combine the partial edges in the subchunks. This gives us a reduced edge list $G_{\rm res}$ which is much smaller than the original edge list $G$. We apply \rm{AGGLOMERATE\_CHUNK} to finish the agglomeration within the original chunk and return the combined dendrogram and the list of the remaining frozen edges.

\begin{algorithm}
    \caption{Distributed clustering agglomerate for large dataset}\label{algorithm:dist}
    \begin{algorithmic}[1]
        \Procedure{\rm{AGGLOMERATE\_RECURSIVE}}{$G, B, T$}
        \If{$G$ is small enough}
            \State $D, G_F \gets$ AGGLOMERATE\_CHUNK$\left( G, B, T \right)$
            \State \textbf{return} $D, G_F$
        \EndIf

        \State $G_{\rm res} \gets \emptyset$
        \State $D \gets \emptyset$
        \State $C \gets$ DIVIDE\_CHUNK$\left( G, B, T \right)$
        \For{$G^C, B^C$ in $C$}
        \State $D^C, G^C_F \gets$ AGGLOMERATE\_RECURSIVE$\left( G^C, B^C, T \right)$ \label{line:upstream}
        \State $G_{\rm res} \gets$ COMBINE\_EDGES$\left(G_{\rm res}, G^C_F\right)$
        \State $D \gets D\cup D^C$
        \EndFor
        \State $D_{\rm res}, G_F \gets$ AGGLOMERATE\_CHUNK$\left( G_{\rm res}, B, T \right)$\label{line:downstream}
        \State $D \gets D\cup D_{\rm res}$
        \State \textbf{return} $D, G_F$
        \EndProcedure
    \end{algorithmic}
\end{algorithm}

The algorithm described in Algorithm \ref{algorithm:dist} allows us to distribute the agglomeration workload to multiple computers. With a shared storage to communicate the inputs and outputs, each \rm{AGGLOMERATE\_CHUNK} call can run on a separate computer. We can design the function \rm{DIVIDE\_CHUNK} and \rm{COMBINE\_EDGES} most convenient for the dataset and cluster. The output does not depend on the detailed implementation of these functions.

There are non-trivial dependencies between the tasks, as one can see the \rm{AGGLOMERATE\_CHUNK} call in line \ref{line:downstream} must wait until the calls in line \ref{line:upstream} have finished. The memory needed to perform the agglomeration task can also vary. It mainly depends on the size of $G_{\rm res}$, which is determined by the number of frozen edges in the subchunks. Intuitively as we stitch bigger and bigger chunks, the edges we need to process in $G_{\rm res}$ will also increase. There may be also big objects that span a large portion of the dataset. Before the chunks grow large enough to contain entire objects, most of the edges related to them will be frozen, potentially contributing significantly to the memory increase.

As long as we use linkage criteria satisfying the reducibility condition, the generic clustering algorithm \ref{algorithm:generic} and the distributed algorithm \ref{algorithm:dist} are equivalent. We can test linkage criteria with small test inputs using Algorithm \ref{algorithm:generic}, and applying it to large datasets using Algorithm \ref{algorithm:dist}

\section{Applications}
\begin{figure}[htpb]
    \centering
    \includegraphics[width=9cm]{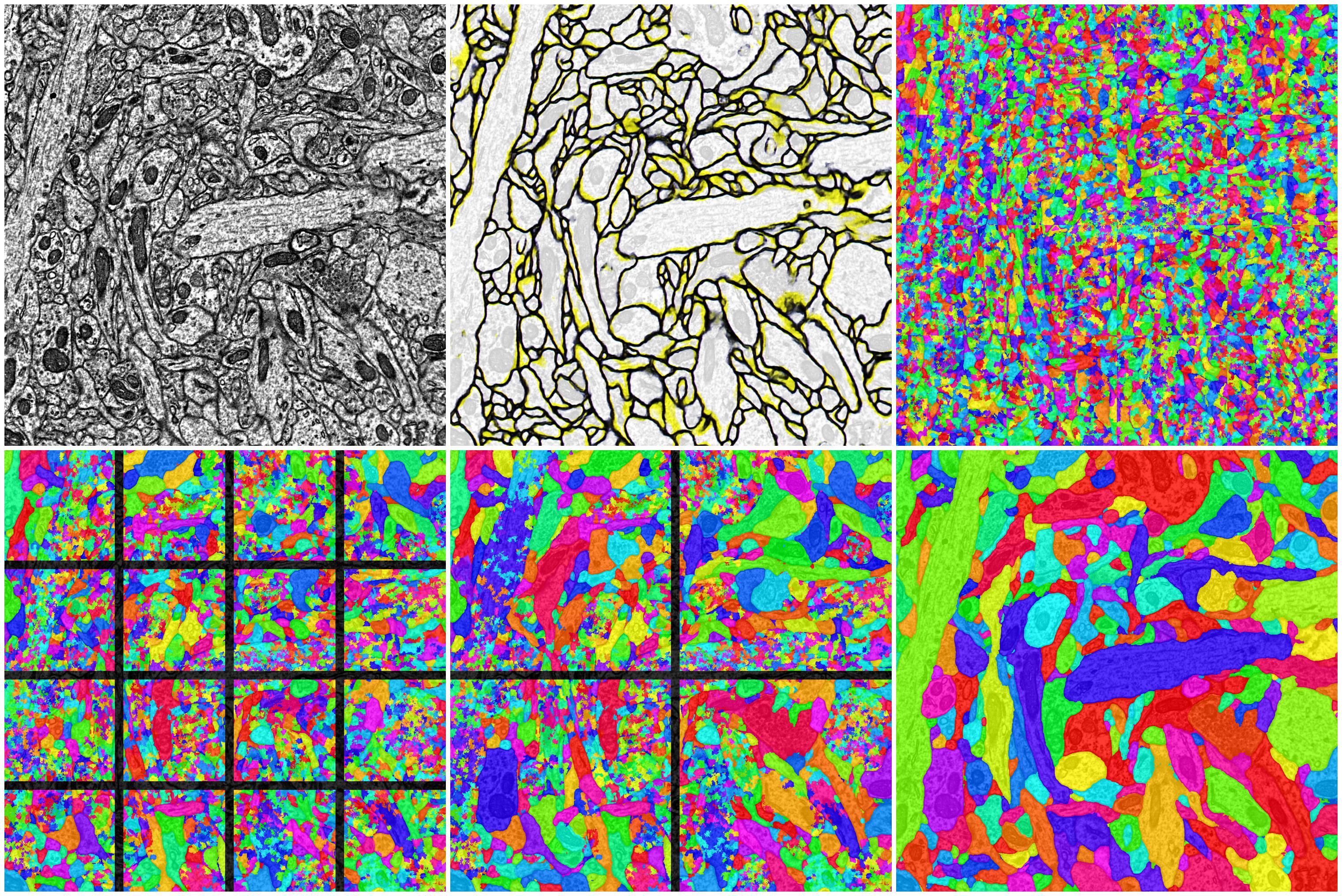} \\
    \vspace{5mm}
    \hspace{-10mm}\includegraphics[width=8cm]{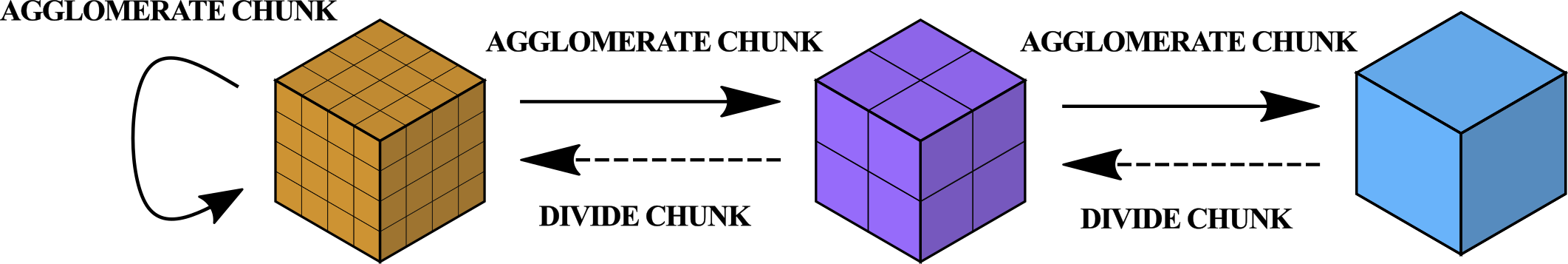}
    \caption{The top panel is an example of the input images and various intermediate outputs of our segmentation pipeline: Starting from an image stack, we first create the nearest neighbor affinity maps using a convolutional neural network. The supervoxel layer is generated by a distributed version of the algorithm described in \citep{ZlateskiThesis}. In the second row, we show how the supervoxels are agglomerated in small chunks and then stitched together. The black lines represent the chunk boundaries. One can clearly see that as the chunks grow, more and more structures are agglomerated, but the supervoxels near the chunk boundaries that may belong to objects across several chunks remain frozen. The panel at the bottom is a schematic diagram of the octree approach. The input image stack is divide into smaller and smaller chunks until they are suitable to be distributed to clusters and cloud computing resources, we then agglomerate the chunks and stitch them back together hierarchically using Algorithm \ref{algorithm:chunked}.}
    \label{fig:pipeline}
\end{figure}

We implemented a system to perform the distributed clustering algorithm we outlined in the previous section. We choose to recursively subdivide the input dataset into eight octants and represent the whole structure as an octree (lower panel of Figure \ref{fig:pipeline}). This allows us to identify boundary supervoxels based on their spatial locations. We use Apache Airflow to keep track of the dependencies between the tasks and assign tasks to different types of computers based on the memory requirement. The upper panel of Figure \ref{fig:pipeline} is an example showing how the system processes an image stack. Starting from the EM images, the system generates the affinity maps and supervoxels needed for the agglomeration process. We first agglomerate in small chunks where many supervoxels have to be frozen due to the boundaries. Gradually we stitch the chunks and agglomerate larger and larger objects. The agglomeration process completes when we stitch all the boundaries together. At this point, we only need to agglomerate supervoxels around the remaining boundaries or those frozen due to big objects. Comparing with the naive approach that agglomerates the entire supervoxel input in one pass, the computation resources required are significantly reduced. Algorithm \ref{algorithm:chunked} is implemented as part of \url{https://github.com/seung-lab/abiss}. And the task orchestration system based on Airflow is available at \url{https://github.com/seung-lab/seuron}.

We tested the system with the Phase I dataset of the MICrONS project \citep{Turner2020.10.14.338681}. It is a $250\times140\times90\; \mu m^3$ region in a mouse visual cortex imaged at $3.58\times3.58\times40\; nm^3$ resolution. The input RAG contains 133 million supervoxels and 1 billion edges between them. We segmented it using Algorithm \ref{algorithm:generic} with the mean affinity linkage criterion. The clustering process took 3 days and required 300 GB of memory. After switching to the system based on Algorithm \ref{algorithm:dist}, the input was organized into a 7-layer octree to distribute the work to up to 100 n1-highmem-32 instances using Google Compute Engine, The same RAG was agglomerated within 2 hours. The leaf nodes are $512 \times 512 \times 128$-voxels chunks. Only 5\% of all the merge operations happened in the top node of the octree. The Phase I results of the MICrONS project can be find in \url{https://microns-explorer.org/}.

In Phase II of the MICrONS project, we segmented a region roughly 1000 times larger. We managed to run the agglomeration process in two weeks with up to 3000 instances using Google Compute Engine. The input image stack has a bounding box of $1.5 \times 1.0 \times 0.5\;  mm^3$ with voxel resolution at $8\times 8 \times 40\; nm^3$. We started from an oversegmentation created by a conservative watershed procedure. The input contained 135 billion supervoxels, the RAG has 1.5 trillion edges. Here we used a 12-layer octree which contained 10 million leaf nodes each represent chunks with up to $256 \times 256 \times 512$ voxels.

\section{Conclusion and Discussion}
\begin{figure}[htpb]
    \centering
    \includegraphics[width=7cm]{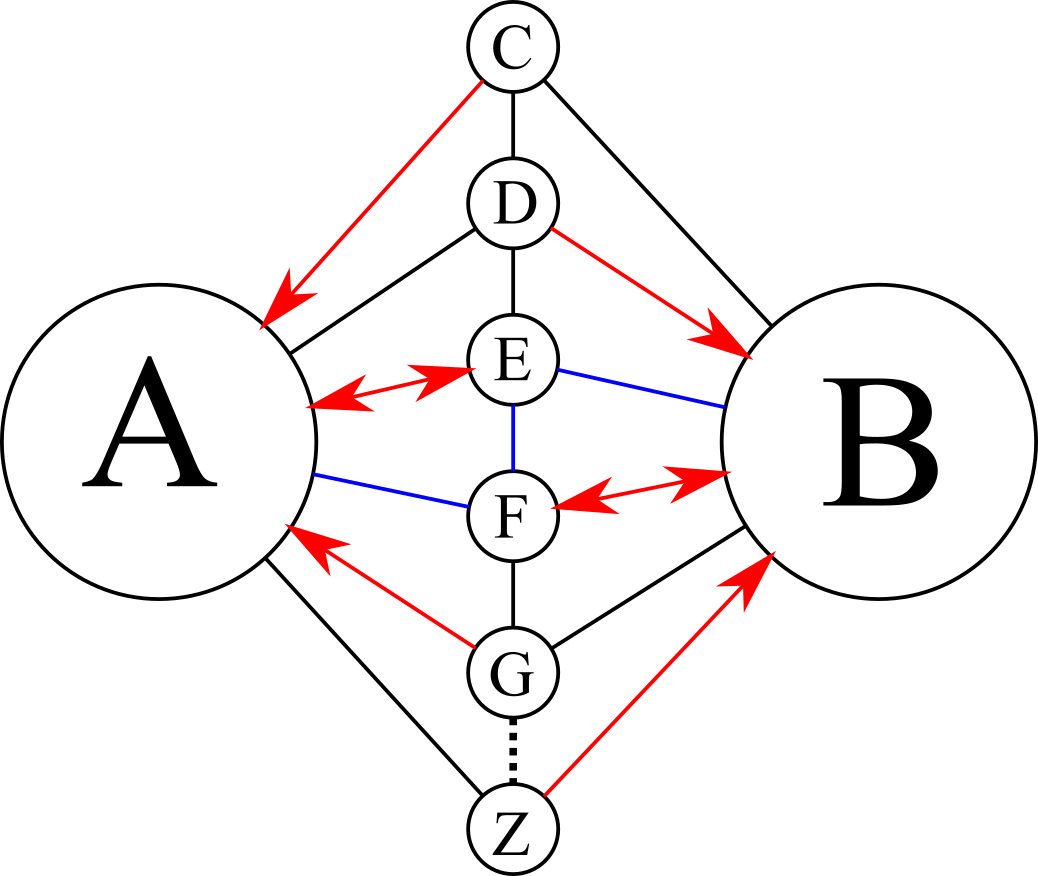}
    \caption{A local structure of RAG common for chunks near the top of the octree. Fragments $A$ and $B$ represents large objects like almost completed neurons. Fragments $C$ to $Z$ are small fragments must be agglomerated to either $A$ or $B$.}
    \label{fig:neartop}
\end{figure}

In this paper we present a framework of distributed clustering algorithms and demonstrated the image segmentation system we built based on it. This framework allows us to agglomerate huge image stacks in chunks to utilize clusters or cloud resources to speed up the process, and avoiding introducing artifacts when we stitch the results together. We make a few remarks before concluding the paper:

a) There are alternative approaches to segment EM images. For example, Flood filling network (FFN) \citep{46992} played a central role in segmenting a portion of the central brain of the fruit fly \citep{Xu2020.01.21.911859}. In principle one can apply FFN  to EM image stack without dividing it into chunks. For the fruit fly, the segmentation are first created in chunks by FFN then merged together conservatively based on the segments in the overlapping regions. While suppressing spurious mergers significantly, the procedure also introduced extra splits. Whether our methods can be adapted to help these approaches is a very interesting open question.

b) We started with a generic clustering algorithm (Algorithm \ref{algorithm:generic}) and demonstrated how to adapt it into a distributed algorithm when the linkage criterion satisfies the reducibility condition with minimal changes. One can also modify nearest-neighbor chain algorithms to consider the chunk boundaries in a similar fashion. In the literature, there are also parallel clustering algorithms based on mutual nearest neighbors that might speed up the process further \citep{1985mca..book.....M}. We tested these implementations in our system, and found they were significant slower than Algorithm \ref{algorithm:chunked} when stitching large chunks near the top of the octree. We believe that several factors contribute to this observation. Algorithm \ref{algorithm:generic} has a complexity $O\left( |E|\log |E| \right)$, while the complexity of a nearest neighbor chain implementation is roughly $O\left( |V| f\left( V \right) \right)$, here $O\left(  f\left( V \right)\right)$ is the complexity of finding the nearest neighbor for a vertex. Because the RAG is sparse\footnote{For example, if we only consider direct contacting supervoxels, the RAG of a 2D image is a planar graph, which satisfy $|E| \le 3 |V| - 6$. For 3D image stack empirically we always have $|E| \le 10 |V|$}, $|E| \propto |V|$. If $O\left( f\left( V \right) \right)$ is a large constant the nearest neighbor chain algorithm will not be efficient. Due to the way we distributing the tasks, the segments remaining incomplete are often large objects surrounded by many small fragments. Thus for the chunks near the top of the octree, a significant portion of the residual RAG have structures look like Figure \ref{fig:neartop}: Segment $A$ and $B$ are two big structure truncated by the chunk boundary, and the rest of the segments $C$-$Z$ are small fragments will be absorb into $A$ or $B$. One can easily see, the nearest neighbor chain construction requires we searching for the nearest neighbor of $A$ or $B$ repeatedly. In our experiment of segmenting large EM image of neural tissue, segments like $A$ and $B$ commonly have several hundreds or even thousands of neighbors. For a brute-force search, $O\left( f\left( V \right) \right)$ is forbiddingly large. Further more, we notice in Figure \ref{fig:neartop}, segment $A, E$ and $B, F$ cannot be agglomerated in parallel without some careful locking mechanism, otherwise there will be a race condition when they both try to merge the blue edges.

c) So far we assumed we divided the input image into non-overlapping chunks. It is also possible to use overlapping chunks in Algorithm \ref{algorithm:dist}. The main advantage is that we can cluster more supervoxels in each step: Any supervoxel will have a FOV at least as large as the overlapping region in at least one chunk. Stitching is possible as long as we still using Algorithm \ref{algorithm:chunked}. The key observation is the following: Suppose chunk $C_1, \cdots, C_n$ share an overlapping region which contains a supervoxel $v$. After clustering each chunk, $v$ now belongs to segment $S_1, \cdots, S_n$. We can prove $S_1, \cdots, S_n$ is a totally ordered set when they are ordered by the dendrogram. Based on this statement, to stitch the overlapping chunks, we need to pick the most advanced segment for each supervoxels in the overlapping region. Overlapping chunks can significantly reduce clustering operations pending until the last few steps, making it possible to cluster much larger datasets comparing to the non-overlapping approach. Of course, because we have to agglomerate the overlapping region several times and the overhead can be significant and undesirable for small input.
\begin{figure}[htpb]
    \centering
    \includegraphics[width=4cm]{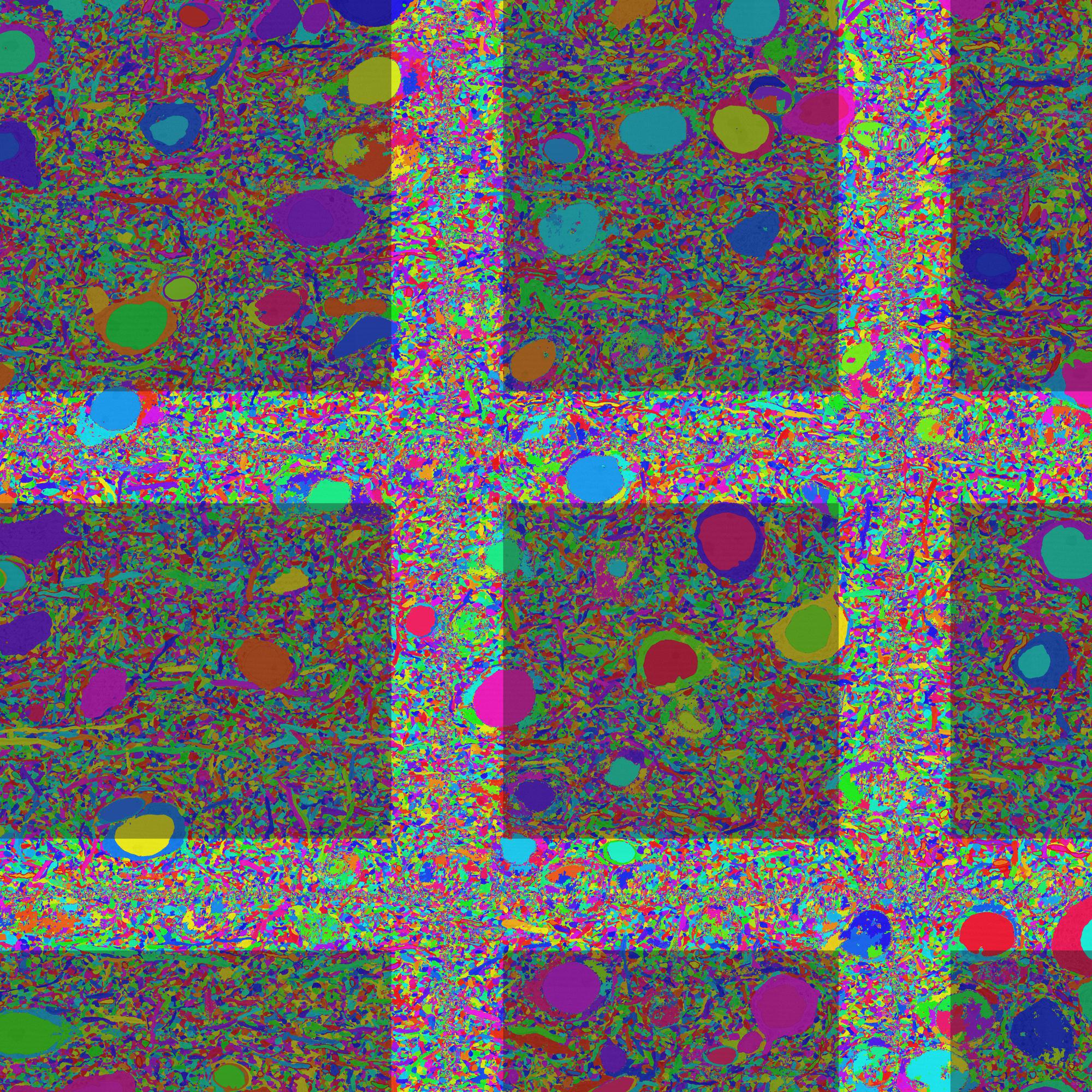}\quad
    \includegraphics[width=4cm]{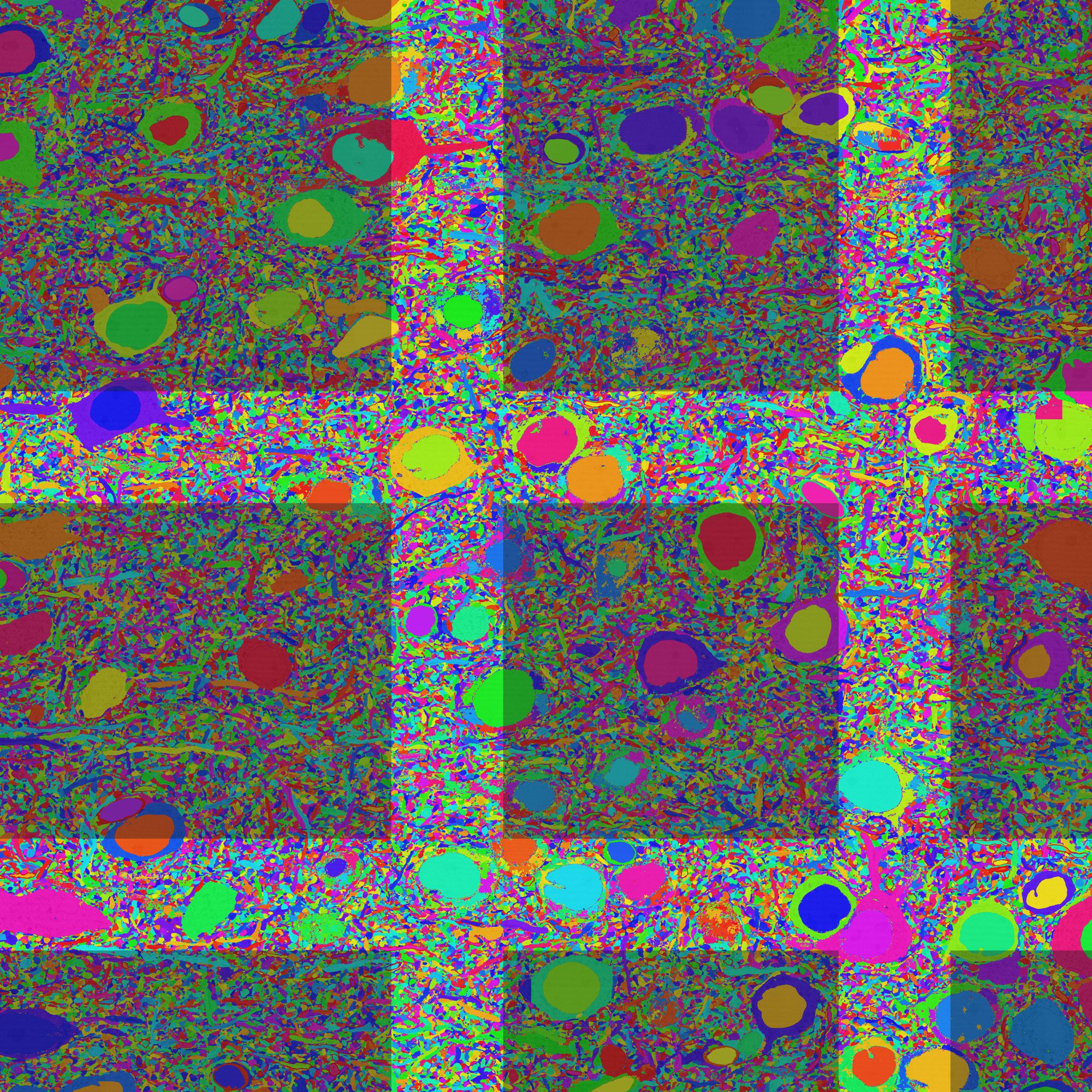}
    \caption{An example of the overlapping chunks reducing the number of fragments needed to be processed during stitching. The dimension of image is $20000\times20000\times512$ voxels, we agglomerate with a 6 layer octree with leaf nodes of $512\times512\times128$ voxels. The figures are the intermediate results at the fourth layer. On the left is the stitched segmentation from non-overlapping $8192\times8192\times512$ chunks. On the right is that from chunks with $2048\times2048\times512$ overlap. The highlighted regions are where the overlaps located. One can clearly see the fragments are reduced in these regions. The final segmentations are still identical up to a relabeling of the segment ids.}
    \label{fig:overlap}
\end{figure}

\section*{Acknowledgments}
We thank Kisuk Lee, Nick Turner, Nico Kemnitz, Will Wong, and Will Silversmith for useful discussions. This research was supported by the Intelligence Advanced Research Projects Activity (IARPA) via Department of Interior/ Interior Business Center (DoI/IBC) contract number D16PC0005, NIH/NIMH (U01MH114824, U01MH117072, RF1MH117815), NIH/NINDS (U19NS104648, R01NS104926), NIH/NEI (R01EY027036), and ARO (W911NF-12-1-0594). The U.S. Government is authorized to reproduce and distribute reprints for Governmental purposes notwithstanding any copyright annotation thereon. Disclaimer: The views and conclusions contained herein are those of the authors and should not be interpreted as necessarily representing the official policies or endorsements, either expressed or implied, of IARPA, DoI/IBC, or the U.S. Government. We are grateful for assistance from Google, Amazon, and Intel.

\bibliographystyle{abbrvnat}
\bibliography{draft}
\end{document}